\documentclass[review]{elsarticle}

\usepackage{xcolor}
\usepackage{graphicx}
\usepackage{amsmath,amssymb}
\usepackage{subfig}
\usepackage{slashbox}
\usepackage{multirow}
\usepackage{lscape}
\usepackage[ruled]{algorithm2e}

\SetKwInput{KwInput}{Input}                
\SetKwInput{KwOutput}{Output}              

\journal{Applied Soft Computing}

\begin{document}

\begin{frontmatter}

\title{LAVARNET: Neural Network Modeling of Causal Variable Relationships for Multivariate Time Series Forecasting}
\author{Christos Koutlis\corref{cor1}}
\ead{ckoutlis@iti.gr}
\author{Symeon Papadopoulos}
\ead{papadop@iti.gr}
\author{Manos Schinas}
\ead{manosetro@iti.gr}
\author{Ioannis Kompatsiaris} 
\ead{ikom@iti.gr}

\cortext[cor1]{Corresponding author}
\address{Information Technologies Institute, Centre of Research and Technology Hellas, Thessaloniki, Greece}

\begin{abstract}
Multivariate time series forecasting is of great importance to many scientific disciplines and industrial sectors. The evolution of a multivariate time series depends on the dynamics of its variables and the connectivity network of causal interrelationships among them. Most of the existing time series models do not account for the causal effects among the system's variables and even if they do they rely just on determining the between-variables causality network. Knowing the structure of such a complex network and even more specifically knowing the exact lagged variables that contribute to the underlying process is crucial for the task of multivariate time series forecasting. The latter is a rather unexplored source of information to leverage. In this direction, here a novel neural network-based architecture is proposed, termed LAgged VAriable Representation NETwork (LAVARNET), which intrinsically estimates the importance of lagged variables and combines high dimensional latent representations of them to predict future values of time series. Our model is compared with other baseline and state of the art neural network architectures on one simulated data set and four real data sets from meteorology, music, solar activity, and finance areas. The proposed architecture outperforms the competitive architectures in most of the experiments.
\end{abstract}


\end{frontmatter}

\section{Introduction}
Time series forecasting is a research topic that attracts great interest in many areas such as meteorology \cite{papacharalampous18}, finance \cite{sezer20,henrique19}, seismology \cite{dibello96}, energy consumption \cite{guo19,jana20}, and traffic \cite{smith02}. Adequately modeling the evolution patterns of time series and thus making accurate estimations of their future values can provide us with crucial information such as warnings about an upcoming storm, earthquake, stock price increase, or traffic jam. Besides, not only the time dependence between past and future values is vital for a system's evolution but also the causal interrelationships among its coupled variables, which might occur in a non-uniform manner \cite{faes12,kugiumtzis13,koutlis19}. Other studies have also pinpointed the importance of non-uniform embeddings of multivariate time series in forecasting \cite{okuno20}.

One of the most well known and widely adopted time series forecasting models, ARIMA \cite{box70}, captures linear correlations between past and future values. Also, many other machine learning methodologies have been employed to serve the same purpose, such as k nearest neighbors \cite{altman92}, Gaussian processes \cite{yan16}, random forests \cite{tyralis17}, multi-layer perceptrons \cite{hamzacebi09} and deep belief networks \cite{kuremoto14}. However today state of the art results in sequence modeling have been produced by recurrent neural network (RNN) architectures \cite{hewamalage19}, which are known to capture the non-linear time dependence between the predicted future value and the preceding values. The standard form of a recurrent neural network is proposed by Elman in \cite{elman90} as in Equation~\ref{eq:rnn}:
\begin{subequations}
\label{eq:rnn}
\begin{align}
h_t&=\sigma_h(W_h x_t+U_h h_{t-1}+b_h)\\
y_t&=\sigma_y(W_y h_t+b_y)
\end{align}
\end{subequations}
where $x_t$ is the input time series at time $t$, $h_t$ is the hidden state of the network at time $t$, $y_t$ is the network’s output at time $t$ and $W_h$, $U_h$, $b_h$, $W_y$, $b_y$ are trainable variables. Also, $\sigma_h$ and $\sigma_y$ are non-linear activation functions.

More complex and effective architectures, both in natural language processing and in time series forecasting\footnote{These two disciplines are subdivisions of the more general term \textit{sequence modeling} and many related methodologies can be applied to both.}, have been proposed since then. Cho et al. \cite{cho14} proposed the encoder-decoder architecture in which an RNN is used to encode the input sequence and a second RNN is then used to decode the encoder's output and make the final prediction. In \cite{bahdanau14} the latter idea is enhanced by an attention mechanism between the encoder and the decoder, which forces the decoder to focus on the most relevant time steps of the input sequence. Some alternative attention mechanisms have also been proposed during the past few years \cite{cinar17,cinar18,shih19}. Other network architectures based on RNNs and Long Short-Term Memory networks (LSTM) \cite{hochreiter97} have been deployed considering both time directions on the input data \cite{schuster97,du20}, skip connections between layers \cite{graves13}, autoregressive components \cite{chang18,lai18}, multi-level attention mechanisms \cite{liang18} and missing values handling \cite{tang19,cinar18}. Also, convolutional neural network (CNN) architectures \cite{borovykh18,koprinska18} and architectures that combine CNNs with other models \cite{sadaei19,cirstea18} have been proposed for regression and forecasting of time series.

However, while all of these architectures capitalize on the estimation of time dependence and global information extraction, none of them accounts for the causal interdependence among the coupled variables of the underlying multivariate mechanism. Recent studies though paved the way towards this direction by employing dual-stage attention mechanisms \cite{qin17,liu20}, which apply attention weights on the input variables during the encoding phase and then apply attention weights on the time steps during the decoding phase. We consider the dual-stage attention-based recurrent neural network (DARNN) \cite{qin17} as a state of the art model in our comparative study presented in the results section.

To the best of our knowledge, although the literature numbers loads of forecasting methodologies none of them takes into account the importance of certain lagged variables in the system's evolution mechanism. Notwithstanding, it is reasonable to consider that extracting more fine-grained information can lead to better forecasting accuracy. For instance, if one variable affects the target after $\tau$ time steps and another variable after $\tau+h$ time steps, knowing only that these two variables affect the target (or even not knowing it) is probably less helpful than being aware of the lags as well.

In our approach, hidden states are generated by the model as high dimensional latent representations of the multivariate time series’ lagged variables, for each pair of variable and time step and not just for each time step. Then, trainable weights are applied to the representations which ideally will tend to foster the correct lagged variables and enable the model to mine this very wanted knowledge of coupling structure. To this end, we changed Elman’s equations by introducing also the variable information in the model in addition to the time step information. Three model versions are proposed here, one that does not consider previous hidden states and consequently being non-recurrent, termed LAgged VAriable Representation NETwork (LAVARNET), one that considers the previous hidden state of the corresponding variable, termed Recurrent LAgged VAriable Representation NETwork (R-LAVARNET) and one that considers the previous hidden states of all variables, termed Fully Recurrent LAgged VAriable Representation NETwork (FR-LAVARNET). 

We conducted a series of experiments using one simulated data set from the well-known difference equations system coupled H\'enon maps \cite{henon76,siggiridou19} and four real data sets from meteorology, music, solar activity, and finance areas. The results show that the proposed architecture is capable of making multivariate and univariate forecasts based on multivariate input signals with great accuracy. Additionally, it outperforms other baseline and state of the art neural network architectures in most of the experiments. Also, a second simulation study reveals the interpretable nature of our model, in which it is shown that the lagged variables that contribute to the system's evolution are fostered by the corresponding trainable weights. Finally, the authors consider as the main contributions of this paper the following:

\begin{itemize}
    
    \item A novel neural network-based architecture is proposed for multivariate time series modeling and forecasting. It considers multivariate input and either multivariate or univariate output depending on the under study problem
    
    \item The main advantage of this method is that estimates the underlying coupling structure among the measured variables of the system and exploits the information gained by the most important lagged variables. Thus, its behavior is interpretable providing extra information regarding the underlying mechanism as the weights of lagged variables estimated at training phase reveal the causal relationships among the measured variables
    
    \item Until now most forecasting methods do not account for the causality patterns among the measured variables and if they do they estimate patterns at variable granularity level. Our method estimates even more fine-grained causality patterns at lagged variable granularity level for the first time
    
    \item This architecture is found to exhibit superior forecasting behavior, compared to other baseline and state of the art models, on one simulated data set and on three out of four real data sets considered in this study
    
\end{itemize}

The rest of the paper is structured as follows. In Section~\ref{sec:methodology} the problem formulation and the proposed architecture are presented, in Section~\ref{sec:data} the data sets are described, in Section~\ref{sec:experiments} the experiments are elaborated and in Section~\ref{sec:conclusions} conclusions are given.

\section{Methodology}\label{sec:methodology}
\subsection{Problem formulation}
The problem of multivariate time series forecasting is formulated as follows. Consider a series of measured signals, \textbf{X}=$[x_{1,:},x_{2,:}\dots,x_{T,:}]$ with $x_{i,:}\in\mathbb{R}^K$ $i=1,2,\dots,T$, where $K$ is the number of variables and $T$ is the number of time steps. The goal is to predict $x_{T+1,:}$ if all variables are of interest or $x_{T+1,k}$ if only variable $k$ is of interest, given \textbf{X}.

In our approach, the estimation of the importance of lagged variables in predicting the future is vital so we would like to clarify the meaning of this notion. Considering a process $x(t)$ at time $t\in\mathbb{N}$, its lagged variables are defined as $x(t-\tau)$, where $\tau\in\mathbb{N}$ is the, so called, lag. In the multivariate case for instance, if $x_{T+1,k}$ is caused by $x_{T,k}$ and $x_{T-1,k-1}$, then variable $k$ at lag $\tau$=1 (namely $x_k(t-1)\equiv x_{T,k}$) and variable $k-1$ at lag $\tau$=2 (namely $x_{k-1}(t-2)\equiv x_{T-1,k-1}$) are the responsible lagged variables for the evolution of variable $k$.

\subsection{LAVARNET: Lagged Variable Representation Network}
Here, a time series forecasting model is proposed that is based on Elman's equations for the recurrent neural network (Equation~\ref{eq:rnn}). The drawback of recurrent neural network architectures that we alleviate here is that they do not account for interrelationships among the time series' variables and hence lack knowledge regarding the underlying causality network of the system. The causal relationships among variables of a coupled multivariate system determine its evolution (along with other factors such as self-dependencies of each variable), making this information crucial for the task of forecasting. To address this problem we add to Elman's equations a term that holds the variable information in addition to the term that holds the time step information when the hidden states are generated. This procedure produces a larger number of hidden states $T\cdot K$ (where $T$ is the number of time steps and $K$ is the number of variables) than all classic recurrent neural networks, that produce $T$ hidden states. On one hand, this requires more memory but on the other hand, additional useful information is leveraged.

As we have already mentioned three model versions are proposed, one that does not consider previous hidden states and consequently can be considered as non-recurrent, termed LAgged VAriable Representation NETwork (LAVARNET), one that considers the previous hidden state of the corresponding variable, termed Recurrent LAgged VAriable Representation NETwork (R-LAVARNET) and one that considers the previous hidden states of all variables, termed Fully Recurrent LAgged VAriable Representation NETwork (FR-LAVARNET).

For the definition of the proposed model, we begin by determining the equations for the hidden states' generation. The LAVARNET's equations are:
\begin{subequations}
\label{eq:lavarnet}
\begin{align}
h_{t,k}&=\sigma_h(W_T\cdot x_{t,:}+W_V\cdot x_{:,k}+b_h)\\
y_{t,k}&=\sigma_y(W_y\cdot h_{t,k}+b_y)
\end{align}
\end{subequations}
The R-LAVARNET's equations are:
\begin{subequations}
\label{eq:rlavarnet}
\begin{align}
h_{t,k}&=\sigma_h(W_T\cdot x_{t,:}+W_V\cdot x_{:,k}+U_h\cdot h_{t-1,k}+b_h)\\
y_{t,k}&=\sigma_y(W_y\cdot h_{t,k}+b_y)
\end{align}
\end{subequations}
Finally, the FR-LAVARNET’s equations are:
\begin{subequations}
\label{eq:frlavarnet}
\begin{align}
h_{t,k}&=\sigma_h(W_T\cdot x_{t,:}+W_V\cdot x_{:,k}+\tilde{U}_h\cdot h_{t-1,:}+b_h)\\
y_{t,k}&=\sigma_y(W_y\cdot h_{t,k}+b_y)
\end{align}
\end{subequations}
where $x_{t,:}\in\mathbb{R}^K$ is the multi-variate input at time $t$, $x_{:,k}\in\mathbb{R}^T$ is the multi-time-step input of variable $k$, $h_{t,k}\in\mathbb{R}^n$ is the hidden state for variable $k$ at time $t$ with $n$ being the number of neurons, $y_{t,k}\in\mathbb{R}^n$ is the output vector for variable $k$ at time $t$, $h_{t-1,:}\in\mathbb{R}^{n\cdot K}$ is the concatenation of all hidden states (all variables) of the previous time step and $W_T$ ($n\times K$), $W_V$ ($n\times T$), $U_h$ ($n\times n$), $\tilde{U}_h$ ($n\times n\cdot K$), $b_h$ ($n\times 1$), $W_y$ ($n\times n$), $b_y$ ($n\times 1$) are trainable variables. Also, $\sigma_h$ and $\sigma_y$ are non-linear activation functions. More specifically here the sigmoid activation is used for both $\sigma_h$ and $\sigma_y$. Then, a matrix of trainable weights $A=\{a_{t,k}\}$ is defined, with $t=1,\dots,T$ and $k=1,\dots,K$ and each scalar $a_{t,k}$ is multiplied with the corresponding output vector $y_{t,k}$. The weights corresponding to important lagged variables should take non-zero values (either positive or negative) and the weights corresponding to non-important lagged variables should take zero (or as close as possible to zero) values after the training procedure. Finally, all output vectors are concatenated in one vector and passed through dense layers for the prediction of future values of the time series. In the case of predicting just one variable’s future values, there is only one fully connected layer in the network’s output and only one trainable matrix of weights $A$, while in the case of predicting many or all of the system’s variables multiple independent fully connected layers are employed as output layers. Additionally, in the latter case, multiple trainable matrices of weights $A_1$, $\dots$, $A_K$ (with $A_i=\{a^i_{t,k}\}$ and $i=1,\dots,K$) are considered, one for each target $i$. The previous step is of utmost importance as each target might be driven by different lagged variables which should be fostered accordingly.

\begin{figure}
\centering
\includegraphics[width=\textwidth]{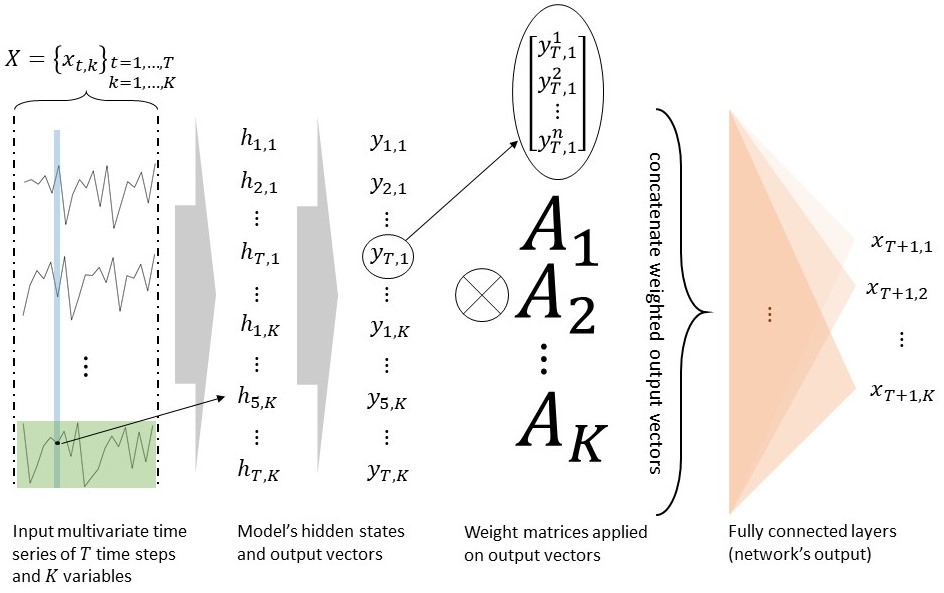}
\caption{Architecture of LAVARNET. Each lagged variable $x_{t,k}$ is represented by a hidden state $h_{t,k}$ which is then transformed to the output vector $y_{t,k}$. Consequently, trainable weights are applied on all output vectors and finally dense layers are employed for the forecasts.}
\label{fig:lavarnet}
\end{figure}

In Figure~\ref{fig:lavarnet}, a graphical representation of our model is illustrated for better comprehension. Each lagged variable $x_{t,k}$ is transformed into the hidden state $h_{t,k}$ through the Equations \ref{eq:lavarnet}a (LAVARNET), \ref{eq:rlavarnet}a (R-LAVARNET) or \ref{eq:frlavarnet}a (FR-LAVARNET) and then the hidden state $h_{t,k}$ is transformed into the output vector $y_{t,k}$ through the Equations \ref{eq:lavarnet}b (LAVARNET), \ref{eq:rlavarnet}b (R-LAVARNET) or \ref{eq:frlavarnet}b (FR-LAVARNET). For the prediction of the first variable ($k=1$) each of the $T\cdot K$ output vectors $y_{t,k}$ is multiplied by the corresponding element $a^1_{t,k}$ of the trainable matrix $A_1$ and consequently, all these new vectors are concatenated in one vector of $T\cdot K\cdot n$ elements, where $n$ is the user-defined number of neurons. The prediction for $x_{T+1,1}$ is then calculated by a fully connected layer. The predictions $x_{T+1,k}$ for the rest variables $k=2,\dots,K$ are performed accordingly using a different matrix $A_k$ and different fully connected layer (network's output) for each variable $k$.

As one may notice, although before the fully connected layers the model exploits all time steps up to time step $T$, the network’s output concerns time step $T+1$, hence there is no information leakage from future to past. The main reason that the proposed architecture performs that well in forecasting is that it is able to capture the connectivity structure of the underlying complex mechanism that generates the measurements. Moreover, it is able to estimate accurately not only the subsystems that contribute to the evolution of each system variable but also the exact lagged variables. This fact makes the results (and consequently the proposed model's behavior) interpretable which is a major advantage and in Section~\ref{subsec:interpretability} a simulation study is presented to showcase this. Finally, we selected as starting point the simple RNNs equations, because any other ideally preferable choice such as Gated Recurrent Unit (GRU) \cite{cho14} or LSTM \cite{hochreiter97} would dramatically increase the number of estimated parameters and thus make the training of the architecture infeasible.

\section{Data}\label{sec:data}
\subsection{Simulated data}\label{subsec:simdata}
The simulated data for the forecasting task are generated by the well-known H\'enon map system \cite{henon76} and more precisely the coupled H\'enon maps system as defined in \cite{siggiridou19} with the chain connectivity pattern among its variables. Many simulation scenarios are considered involving different number of variables $K$=5,10,15, number of time steps $T$=3,5,10,15 and time series lengths $L$=200, 500, 1000, 2000, 3000, 4000, 5000, 6000, 7000, 8000, 9000 and 10000. For each simulation scenario (e.g.  $K$=5, $T$=10 and $L$=1000) 5 Monte-Carlo simulations  are generated in order to obtain average model performance.

For the interpretability simulation study presented in Section~\ref{subsec:interpretability} we use the linear stochastic process of vector autoregressive model (VAR) \cite{basu15} with a random network of connections \cite{erdos59} among its variables for the generation of multivariate time series. The time series length is set to $L$=5000 and different simulation scenarios in terms of number of variables $K$=2,3,…15 and model order $P$=1,2,3 are considered. Also, 10 Monte-Carlo simulations per simulation scenario are generated for the estimation of average model performance.

\subsection{Real data}
For the real data analysis four data sets from different research domains are considered, one data set related to weather (SML2010)\footnote{https://archive.ics.uci.edu/ml/datasets/SML2010}, a set of data related to musical genre popularity (GenrePopularity) that is generated as part of the FuturePulse project\footnote{http://www.futurepulse.eu/}, a data set related to solar activity (Solar-Energy)\footnote{https://www.nrel.gov/grid/solar-power-data.html} and a data set related to finance (Currency)\footnote{https://www.kaggle.com/thebasss/currency-exchange-rates}.

The SML2010 data set contains indoor temperature time series and other relevant quantities like Carbon dioxide in ppm and sunlight in the south facade. This data set is collected from a monitor system mounted on a domestic house. Our target variable is the room temperature and 16 other relevant driving series are used as input to our models as well. The data were sampled every minute and was smoothed with 15-minute means. In this study, we use the first 3200 data points as the training set, the following 400 data points as the validation set, and the last 537 data points as the test set.

The GenrePopularity data set contains time series data from 2000-01-01 until 2019-10-31 related to the popularity level of 60 musical genres (presented in Table~\ref{tab:genre_names}) in four countries: Great Britain, United States of America, Sweden and Canada. Each time series point concerns the percentage of entries, in charts of a certain country, related to a specific musical genre for a sliding time window of 4 weeks with a step of one week. We have collected data from 60 charts in Great Britain, 116 charts in the United States of America, 26 charts in Sweden, and 18 charts in Canada. Then by aggregating the entries, 4 multi-variate time series are generated with 60 variables and 1031 time points each. Also, a moving average filter of order 4 is applied for noise reduction. Training, validation, and test sets are generated by splitting the time series into 618, 206, and 207 time points respectively. Most of the time series variables are sparse, thus for the forecasting task all genre time series with more than 100 zeros are discarded from the models’ input. As target variables, we selected three non-sparse musical genres namely Pop, Rock, and Hip-hop.

\begin{table}
    \footnotesize  
    \centering
    \begin{tabular}{lllll}
        \hline
        African& Alternative& Ambient& Americana& Bass\\
        Blues&Breakbeat& Children's Music& Christian& Christmas\\
        Classical& Country&Dance/EDM& Reggaeton& Death Metal\\
        Disco& Doom Metal& Downbeat&Drum\&Bass& Dubstep\\
        Electronic& Electronica& Experimental& Folk&Funk\\
        Garage& Hardcore& Hard Rock& Heavy Metal& Hiphop\\
        House& Techno& Indie& Industrial& Inspirational\\
        Instrumental&Jazz& Karaoke& Latin& Lounge\\
        Mariachi& Metal&Musical& Opera& Pop\\
        R\&B& Reggae& Rock&Rockabilly& Salsa\\
        Samba& Singer-Songwriter& Soul& Soundtrack&Spoken Word\\
        Surf& Tango& Tech House& Thrash Metal& TripHop\\
        \hline
    \end{tabular}
    \caption{Musical genres considered in the GenrePopularity data set.}
    \label{tab:genre_names}
\end{table}

The Solar-Energy data set contains time series data about the solar power production records in the year of 2006, which is sampled every 10 minutes from 137 photovoltaic plants in Alabama State. The total number of time points is 52,560 and we split it into training, validation, and test sets with 31,536, 10,512 and 10,512 time points respectively. As target variables, we use the first 10 variables, after sorting the file names.

The Currency data set contains the daily currency exchange rates as reported to the International Monetary Fund by the issuing central bank. Included are 51 currencies over the period from 01-01-1995 to 11-04-2018. The format is known as currency units per U.S. Dollar. Explained by example, each rate in the Euro column says how much U.S. Dollar you had to pay at a certain date to buy 1 Euro. Hence, the rates in the column U.S. Dollar are always 1. We use data from 30-10-1998 which is the date Euro was released, currencies with more than 1500 missing values and currencies with constant exchange rate are discarded and the rest are linearly interpolated. So finally, the data set contains 41 currencies and 4,986 time points which are split into 2991 training samples, 997 validation samples, and 998 test samples. As target variables we use the first ten\footnote{We skip the second currency Bahrain Dinar as it exhibits constant exchange rate in the test set.} currencies being Australian Dollar (1), Botswana Pula (2), Brazilian Real (3), Brunei Dollar (4), Canadian Dollar (5), Chilean Peso (6), Chinese Yuan (7), Colombian Peso (8), Czech Koruna (9), Danish Krone (10).

As a pre-processing step z-score normalization is applied to SML2010, Solar-Energy and Currency, while no normalization is applied to GenrePopularity which contains values between 0 and 1 by default.

\section{Experiments}\label{sec:experiments}
\subsection{Competitive models}
In order to perform a comparative study we consider three baseline and two state of the art time series models as competitive models. The three baselines are (1) the k nearest neighbors regression model (KNN), (2) the single layered recurrent neural network (RNN) \cite{elman90} and (3) the single layered long short-term memory network (LSTM) \cite{hochreiter97} each followed by a dense output layer. The two state of the art models are (1) the dual-stage attention-based recurrent neural network (DARNN) \cite{qin17}, which considers an encoder with attention on the input variables and a decoder with attention on the time steps, and (2) the WaveNet \cite{oord16} as proposed in \cite{borovykh17} for time series forecasting, which is mainly a stack of dilated convolutional layers with residual and skip connections.

For KNN, 5 neighbors are used and for RNN and LSTM, 128 neurons are considered in all real data experiments as baseline selections. For DARNN, the optimal number of neurons is determined after grid search among 32, 64, 128 neurons and for LAVARNET grid search among 5, 10, 20\footnote{Except for the Currency data set in which 32, 64 and 128 neurons are selected for the grid search because higher values produced better results in this data set.} is also employed. Finally, for WaveNet 64 filters and filter width 2 are considered for each of 6 stacked dilated convolutional layers with dilation rates 1, 2, 4, 1, 2, 4 respectively.

\subsection{Training}
For the training of all neural network-based models we used Adam optimizer and mean squared error loss function provided by TensorFlow 1.8.0. Also, three GPU devices (two GeForce GTX 1080 and one GeForce GTX 1070) are employed for the experiments. Additionally, for the KNN training the scikit-learn Python package implementation is employed.

In the experiments, the data sets were split into training (60\%), validation (20\%) and test (20\%) sets (except for the SML2010 data set in which we opted for the same splitting as in \cite{qin17} for comparison purposes). So, the models' training is performed on the training set, all models are checkpointed based on their performance on the unseen data of the validation set and their performance is evaluated on the test set.

Also, for DARNN we opt for the proposed in \cite{qin17} learning rate strategy, starting with 0.001 and reducing by 10\% every 10,000 iterations. For RNN, LSTM, and WaveNet we opt for a constant learning rate equal to 0.001 and for LAVARNET cosine annealing \cite{loshchilov17} is used, where at epoch $i$ the learning rate $\eta(i)$ is set to:
\begin{equation}
    \eta(i)=\eta_{min}+\frac{1}{2}(\eta_{max}-\eta_{min})\bigg(1+cos\big(\frac{i\cdot\pi}{E}\big)\bigg)
\end{equation}
where $\eta_{max}$=0.01, $\eta_{min}$=0.0001 and $E$ is the number of epochs.

\subsection{Evaluation}
For the evaluation of all models on the forecasting task we use the error function mean absolute error (MAE) as in Equation~\ref{eq:mae}:
\begin{equation}\label{eq:mae}
    MAE_k=\frac{1}{N}\sum_{t=1}^N \mid x_{t,k}-\hat{x}_{t,k}\mid
\end{equation}
where $k$ is the target variable, $N$ is the number of samples in the test set, $x_{t,k}$ is the actual measurement and $\hat{x}_{t,k}$ is the prediction. In the case of multivariate prediction the average $MAE_k$ across all $k$ is considered.

For the interpretation of the results in terms of correct weighting of lagged variables (Section~\ref{subsec:interpretability}), we use the percentage of true driving lagged variables that were among the ones with the highest absolute weights assigned by LAVARNET. More precisely, in the simulations the exact lagged variables that drive each target variable is known. Let us say $L_k$ is the set of lagged variables that drive target variable $k$. Then, for that target variable we determine the set of lagged variables $\tilde{L_k}$ that are associated with the $C(L_k)$ (where $C(S)$ is the cardinality of set $S$) highest absolute values of matrix $A_k$. Finally, using Equation~\ref{eq:rl}, we compute the success percentage of true driving lagged variables of the whole system that were categorized by LAVARNET as such and use this score as evaluation index:
\begin{equation}\label{eq:rl}
    R_L = \frac{\sum_kC(L_k\cap\tilde{L_k})}{\sum_kC(L_k)}
\end{equation}
where $k$ is the target variable.

Also, we consider another less strict score namely the percentage of true driving variables that were categorized by LAVARNET as such. It is denoted by $R_V$ and computed as in Equation~\ref{eq:rv}:
\begin{equation}\label{eq:rv}
    R_V = \frac{\sum_kC(V_k\cap\tilde{V_k})}{\sum_kC(V_k)}
\end{equation}
where $k$ is the target variable, $V_k$ is the set of variables that drive target variable $k$ and $\tilde{V_k}$ is the set of variables that are associated with the $C(L_k)$ highest absolute values of matrix $A_k$.

\subsection{Results}
\begin{table}
    \centering
    \begin{tabular}{ll}
    \hline
    model &	MAE\\
    \hline\hline
    \textbf{LAVARNET} &	\textbf{0.0430}\\
    R-LAVARNET 	&0.0442\\
    FR- LAVARNET &	0.0460\\
    LSTM &	0.0534\\
    RNN &	0.0561\\
    KNN &	0.1473\\
    \hline
    \end{tabular}
    \caption{Average performance of models on the simulation data set, across all scenarios and Monte-Carlo simulations.}
    \label{tab:simulation}
\end{table}

First, we present results for the simulation study in which our model is compared with the baseline models KNN, single layered RNN and single layered LSTM. In the simulation study we evaluate forecasting models in multivariate prediction, thus DARNN and WaveNet are discarded as not applicable\footnote{DARNN model is designed to make univariate forecasts given multivariate signals as input and WaveNet is disigned to model univariate signals.}. Also, 100 neurons are considered for all neural network-based models' hidden state vectors in the simulation study. As described in Section~\ref{subsec:simdata} many simulation scenarios in terms of number of variables, number of time steps and time series length are considered as well as multiple Monte-Carlo simulations of each scenario. In Table~\ref{tab:simulation}, the average models’ performance across all different scenarios and Monte-Carlo simulations is presented and the model exhibiting the best performance is highlighted with bold letters, being LAVARNET.

\begin{figure}
    \centering
        \subfloat[]{\includegraphics[width=0.5\textwidth]{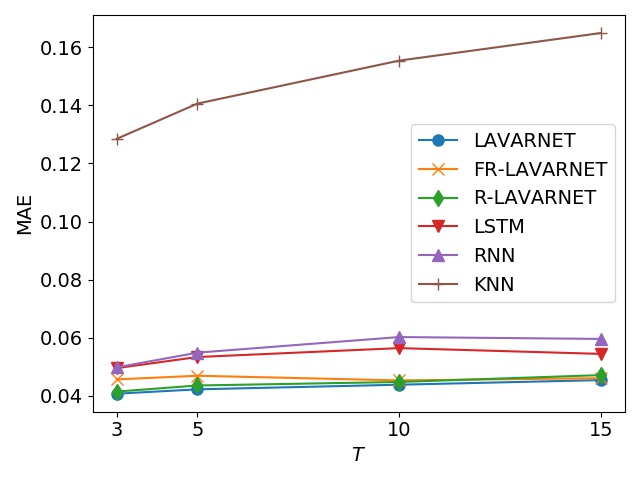}}
        \subfloat[]{\includegraphics[width=0.5\textwidth]{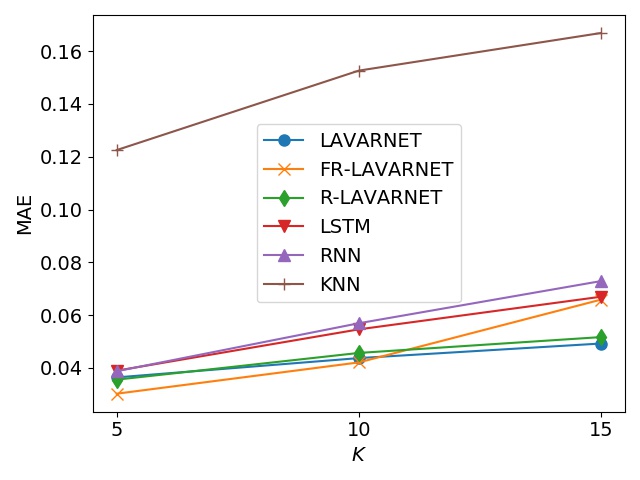}}
        \quad
        \subfloat[]{\includegraphics[width=0.5\textwidth]{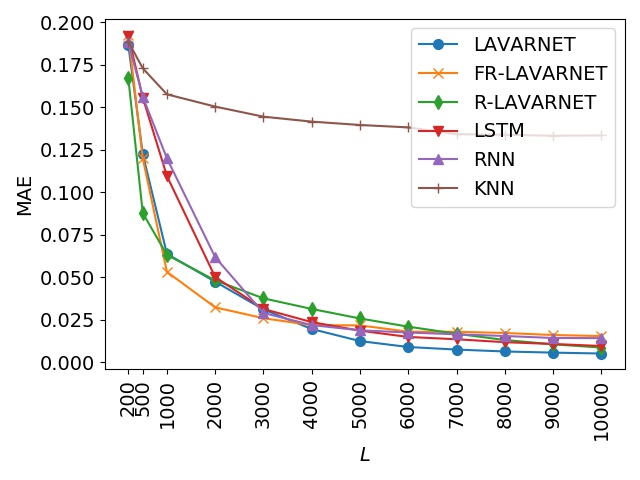}}
    \caption{The performance of LAVARNET, R-LAVARNET, FR-LAVARNET, LSTM, RNN and KNN in terms of mean absolute error (MAE) on multivariate prediction task for the simulated coupled H\'enon maps system. (a) Average performance across number of variables, time-series length and Monte-Carlo simulations, (b) average performance across number of time steps, time-series length and Monte-Carlo simulations and (c) average performance across number of time steps, number of variables and Monte-Carlo simulations.}
    \label{fig:simulation}
\end{figure}

For more details Figure~\ref{fig:simulation} is also provided. In this figure the average performance of the models is illustrated excluding the parameter (a) time steps, (b) number of variables and (c) time series length, respectively. It is observed that all models’ performance decreases as the number of time steps and number of variables increases and that all models’ performance increases as the time series length increases, as expected. The proposed models perform better than the baselines in all scenarios but LAVARNET and R-LAVARNET exhibit a more consistent behavior, especially when the number of variables is high. This is actually expected given that FR-LAVARNET’s number of parameters is rapidly increasing with the number of variables. Additionally, R-LAVARNET produces the smallest errors when the time series length is small, while FR-LAVARNET and LAVARNET perform best when the time series length is of intermediate and large size respectively.

\begin{table}
    \centering
    \begin{tabular}{ll}
    \hline
    model&MAE\\
    \hline\hline
    LAVARNET&	0.0674$\pm$0.012\\
    \textbf{R-LAVARNET}&	\textbf{0.0389}$\pm$\textbf{0.012}\\
    FR-LAVARNET&	0.0804$\pm$0.013\\
    LSTM&	0.0849$\pm$0.007\\
    RNN&	0.0967$\pm$0.010\\
    DARNN&	0.0533$\pm$0.004\\
    WaveNet & 0.0866$\pm$0.0.0253\\
    KNN & 0.4983\\
    \hline
    \end{tabular}
    \caption{Evaluation of models on SML2010 data set using mean absolute error (MAE). The average value of the error function is presented along with the corresponding standard deviation. The model exhibiting the best performance is highlighted with bold letters.}
    \label{tab:sml2010}
\end{table}

In Table~\ref{tab:sml2010}, the results for the comparative study on the real data set SML2010, are presented. 
Apparently, R-LAVARNET produces the smallest errors and although overlapping, the intersection of uncertainty intervals with the second one, being DARNN, is very small. The WaveNet's performance is comparable to the performances of RNN, LSTM and FR-LAVARNET in this experiment and KNN produces the greatest errors.

In Table~\ref{tab:genres}, the performance of forecasting models on the task of predicting musical genres’ popularity is presented. We consider 12 different settings each of them related to a different country and a different target variable (musical genre). Also, the presented results are the average and standard deviation of MAE across 10 repetitions of each training/testing procedure. 
It is observed that in 7 settings LAVARNET performs best, in 2 settings R-LAVARNET performs best, in one setting the simple RNN model performs best and only in 2 settings DARNN outperforms all the other models. WaveNet and KNN yield the least accurate forecasts in most of the settings of this data set, with KNN producing the largest errors.

\begin{table}
    \footnotesize
    \centering
    \begin{tabular}{llllll}
    \hline
    &&GB&	US&	SE&	CA\\
    \hline\hline
    \multirow{6}{*}{\rotatebox[origin=c]{90}{Pop}}&LAVARNET&\textbf{0.0049}$\pm$\textbf{0.0003}&0.0100$\pm$0.0021&0.0087$\pm$0.0016&0.0110$\pm$0.0019\\
    &R-LAVARNET& 0.0060$\pm$0.0003&0.0160$\pm$0.0022&\textbf{0.0075}$\pm$\textbf{0.0005}&0.0100$\pm$0.0023\\
    &FR-LAVARNET& 0.0087$\pm$0.0007&0.0280$\pm$0.0049&0.0083$\pm$0.0008&0.0114$\pm$0.0022\\
    &LSTM*&0.0114$\pm$0.0038&0.0180$\pm$0.007&0.0108$\pm$0.004&0.0096$\pm$0.004\\
    &RNN*&0.0116$\pm$0.0033&0.0312$\pm$0.010&0.0124$\pm$0.006&
    \textbf{0.0089}$\pm$\textbf{0.003}\\
    &DARNN& 0.0092$\pm$0.0005&\textbf{0.0077}$\pm$\textbf{0.0005}&0.0086$\pm$0.0001&0.0114$\pm$0.0008\\
    &WaveNet&0.0174$\pm$0.0127&0.0509$\pm$0.0237&0.0228$\pm$0.0053&0.0175$\pm$0.0076\\
    &KNN&0.0649&0.0761&0.0536&0.0175\\
    \hline
    \multirow{6}{*}{\rotatebox[origin=c]{90}{Rock}}&LAVARNET&0.0068$\pm$0.0004&\textbf{0.0029}$\pm$\textbf{0.0001}&\textbf{0.0017}$\pm$\textbf{0.0003}&0.0030$\pm$0.0005\\
    &R-LAVARNET& 0.0076$\pm$0.0006&0.0034$\pm$0.0004&0.0022$\pm$0.0003&\textbf{0.0028}$\pm$\textbf{0.0005}\\
    &FR-LAVARNET& 0.0103$\pm$0.0007&0.0044$\pm$0.0007&0.0026$\pm$0.0005&0.0048$\pm$0.0020\\
    &LSTM& 0.0114$\pm$0.003&0.0150$\pm$0.005&0.0050$\pm$0.001&0.0050$\pm$0.002\\
    &RNN& 0.0106$\pm$0.002&0.0137$\pm$0.007&0.0074$\pm$0.003&0.0055$\pm$0.002\\
    &DARNN&\textbf{0.0062}$\pm$\textbf{0.0004}&0.0043$\pm$0.0002&0.0026$\pm$0.00009&0.0041$\pm$0.0003\\
    &WaveNet&0.0209$\pm$0.0115&0.0096$\pm$0.0057&0.0197$\pm$0.0331&0.0088$\pm$0.0043\\
    &KNN&0.0276&0.0133&0.0384&0.0116\\
    \hline
    \multirow{6}{*}{\rotatebox[origin=c]{90}{Hip-hop}}&LAVARNET&\textbf{0.0031}$\pm$\textbf{0.0007}&\textbf{0.0039}$\pm$\textbf{0.0003}&\textbf{0.0023}$\pm$\textbf{0.0004}&\textbf{0.0048}$\pm$\textbf{0.0014}\\
    &R-LAVARNET& 0.0040$\pm$0.0004&0.0042$\pm$0.0005&0.0029$\pm$0.0004&0.0065$\pm$0.0011\\
    &FR-LAVARNET&0.0054$\pm$0.0011&0.0065$\pm$0.0012&0.0049$\pm$0.0004&0.0102$\pm$0.0014\\
    &LSTM& 0.0108$\pm$0.0022&0.0131$\pm$0.0042&0.0067$\pm$0.0031&0.0112$\pm$0.0037\\
    &RNN& 0.0151$\pm$0.0044&0.0145$\pm$0.007&0.0063$\pm$0.0028&0.0132$\pm$0.004\\
    &DARNN& 0.0059$\pm$0.0006&0.0064$\pm$0.0020&0.0053$\pm$0.00009&0.0088$\pm$0.0013\\
    &WaveNet&0.0102$\pm$0.0047&0.0159$\pm$0.0080&0.0089$\pm$0.0027&0.0134$\pm$0.0078\\
    &KNN&0.0142&0.0134&0.0146&0.0331\\
    \hline
    \end{tabular}
    \caption{Evaluation of models on musical genre popularity forecasting task for three different musical genres (Pop, Rock, Hip-hop) as target in four different countries (Great Britain; GB, United States; US, Sweden; SE, Canada; CA). The evaluation index is MAE (the standard deviation is also presented) and the model exhibiting best performance at each setting is highlighted with bold letters. 
    }
    \label{tab:genres}
\end{table}

For the Solar-Energy data set that contains time ordered measurements from 137 variables, 10 comparative experiments are conducted forecasting future values of each of the first 10 variables, after sorting the file names. A multivariate prediction experiment would leave DARNN out of the comparative study and also would be infeasible in terms of GPU memory consumption. Additionally, conducting 137 separate experiments is computationally very costly, thus we opted for the first 10 variables. 
In Table~\ref{tab:solar} the corresponding results are illustrated, in which R-LAVARNET performs best in all 10 experiments. LAVARNET and LSTM produce comparable to R-LAVARNET's errors and all the rest models exhibit worse performance in the Solar-Energy data set.

\begin{landscape}
    \begin{table}
        \small
        \centering
        \begin{tabular}{lllllllll}
    \hline
        $k$ &	LAVARNET&	R-LAVARNET&	FR-LAVARNET&	LSTM&	RNN&	DARNN&	WaveNet&	KNN\\
    \hline\hline
        1&	1.18$\pm$0.09&	\textbf{1.13}$\pm$\textbf{0.04}&	1.33$\pm$0.13&	1.40$\pm$0.27&	1.43$\pm$0.24&	1.53$\pm$0.08 & 1.57$\pm$0.23& 2.26\\
        2&	1.30$\pm$0.10&	\textbf{1.26}$\pm$\textbf{0.08}&	1.50$\pm$0.18&	1.38$\pm$0.12&	1.51$\pm$0.18&	1.65$\pm$0.04& 2.06$\pm$0.38& 2.71\\
        3&	0.38$\pm$0.03&	\textbf{0.33}$\pm$\textbf{0.04}&	0.42$\pm$0.05&	0.42$\pm$0.10&	0.44$\pm$0.09&	0.49$\pm$0.02& 0.67$\pm$0.34& 0.92\\
        4&	0.39$\pm$0.03&	\textbf{0.36}$\pm$\textbf{0.06}&	0.44$\pm$0.06&	0.47$\pm$0.09&	0.50$\pm$0.09&	0.54$\pm$0.05&0.64$\pm$0.21& 0.94\\
        5&	0.39$\pm$0.03&	\textbf{0.37}$\pm$\textbf{0.04}&	0.43$\pm$0.06&	0.38$\pm$0.04&	0.45$\pm$0.07&	0.53$\pm$0.03&0.67$\pm$0.16& 0.97\\
        6&	0.87$\pm$0.09&	\textbf{0.83}$\pm$\textbf{0.04}&	0.92$\pm$0.11&	0.97$\pm$0.16&	1.00$\pm$0.10&	1.05$\pm$0.04&1.14$\pm$0.14& 1.60\\
        7&	0.39$\pm$0.04&	\textbf{0.35}$\pm$\textbf{0.04}&	0.41$\pm$0.03&	0.43$\pm$0.09&	0.49$\pm$0.10&	0.49$\pm$0.04&0.56$\pm$0.10& 0.92\\
        8&	0.36$\pm$0.03&	\textbf{0.33}$\pm$\textbf{0.04}&	0.40$\pm$0.07&	0.37$\pm$0.07&	0.39$\pm$0.06&	0.47$\pm$0.03&0.45$\pm$0.05& 0.93\\
        9&	0.38$\pm$0.02&	\textbf{0.34}$\pm$\textbf{0.03}&	0.43$\pm$0.05&	0.37$\pm$0.04&	0.43$\pm$0.08&	0.52$\pm$0.04&0.63$\pm$0.21& 0.98\\
        10&	0.37$\pm$0.03&	\textbf{0.34}$\pm$\textbf{0.04}&	0.44$\pm$0.09&	0.39$\pm$0.04&	0.47$\pm$0.09&	0.51$\pm$0.03&0.63$\pm$0.21& 0.92\\
    \hline
        \end{tabular}
        \caption{Evaluation of models on Solar-Energy data set for the first 10 variables. The evaluation index is MAE (the standard deviation is also presented) and the  model  exhibiting  best  performance  at  each  experiment (target variable $k$)  is  highlighted with  bold  letters. 
        }
        \label{tab:solar}
    \end{table}
\end{landscape}

In Table~\ref{tab:currency}, the results for the Currency data set are presented. For the same reasons as in the previous real data analysis we present results for the first 10 variables.
In the results it is observed that our architectures exhibit the best performance only in 5 out of 10 experiments, while DARNN outperforms the other models in the rest 5 experiments. The baseline models KNN, RNN and LSTM do not outperform the other models in any experiment as expected. Also, the WaveNet outperforms LAVARNET in 3 out of 10 experiments, but in none of them exhibits the best performance across all models.

\begin{landscape}
    \begin{table}
        \small
        \centering
        \begin{tabular}{lllllllll}
    \hline
        $k$ &	LAVARNET&	R-LAVARNET&	FR-LAVARNET&	LSTM&	RNN&	DARNN&	WaveNet&	KNN\\
    \hline\hline
        1&\textbf{0.011}$\pm$\textbf{0.005}&0.018$\pm$0.005&0.018$\pm$0.008&0.047$\pm$0.024&0.040$\pm$0.013&0.024$\pm$0.008&0.014$\pm$0.002& 0.029\\
        2&\textbf{0.008}$\pm$\textbf{0.003}&0.012$\pm$0.006&0.014$\pm$0.004&0.047$\pm$0.028&0.138$\pm$0.096&0.009$\pm$0.004&0.031$\pm$0.020& 0.051\\
        3&\textbf{0.049}$\pm$\textbf{0.007}&0.060$\pm$0.020&0.224$\pm$0.077&0.287$\pm$0.146&0.367$\pm$0.183&0.102$\pm$0.036&0.146$\pm$0.048& 0.376\\
        4&0.069$\pm$0.007&0.092$\pm$0.012&0.049$\pm$0.020&0.126$\pm$0.121&0.069$\pm$0.034&\textbf{0.026}$\pm$\textbf{0.015}&0.033$\pm$0.015& 0.072\\
        5&0.022$\pm$0.004&0.024$\pm$0.006&0.029$\pm$0.011&0.112$\pm$0.063&0.106$\pm$0.066&\textbf{0.017}$\pm$\textbf{0.003}&0.061$\pm$0.048& 0.065\\
        6&13.74$\pm$5.07&\textbf{9.75}$\pm$\textbf{7.72}&11.38$\pm$3.38&66.27$\pm$32.27&25.05$\pm$17.53&10.18$\pm$4.14&18.44$\pm$24.46& 43.35\\
        7&0.330$\pm$0.017&0.361$\pm$0.020&0.338$\pm$0.021&0.348$\pm$0.165&0.353$\pm$0.178&\textbf{0.069}$\pm$\textbf{0.026}&0.23$\pm$0.068& 0.280\\
        8&59.56$\pm$8.62&62.25$\pm$10.26&170.4$\pm$21.6&194.9$\pm$118.2&201.1$\pm$92.33&\textbf{47.43}$\pm$\textbf{17.86}&206.6$\pm$225.3& 180.7\\
        9&1.35$\pm$0.392&1.63$\pm$0.703&1.20$\pm$0.285&9.24$\pm$6.12&3.52$\pm$2.81&\textbf{0.751}$\pm$\textbf{0.097}&0.823$\pm$0.463& 0.982\\
        10&0.095$\pm$0.022&0.076$\pm$0.025&\textbf{0.074}$\pm$\textbf{0.026}&0.822$\pm$0.517&0.499$\pm$0.303&0.104$\pm$0.063&0.130$\pm$0.050& 0.394\\
    \hline
        \end{tabular}
        \caption{Evaluation of models on Currency data set for the first 10 variables. The evaluation index is MAE (the standard deviation is also presented) and the  model  exhibiting  best  performance  at  each  experiment (target variable $k$)  is  highlighted with  bold  letters. 
        }
        \label{tab:currency}
    \end{table}
\end{landscape}

Finally, the fact that FR-LAVARNET does not frequently perform better than LAVARNET and R-LAVARNET has a twofold explanation. The first reason is that FR-LAVARNET involves a much higher number of trainable parameters especially in cases of many coupled variables. Specifically, the matrix $\tilde{U}_h$ of Equation~\ref{eq:frlavarnet} has size $n\times n\cdot K$, while R-LAVARNET’s corresponding matrix $U_h$ has size $n\times n$ and LAVARNET does not even involve such a matrix. Second, FR-LAVARNET is likely to incorporate excessive or irrelevant information through the hidden states of the other variables at time step $t-1$.

\subsection{Interpretability simulation study}\label{subsec:interpretability}
Here, we use the VAR model for the generation of multivariate time series of different dimensions and model orders. The causal relationships among the variables of the system are selected at random using the Erd\"os-R\'enyi random network scheme with 40\% network density and for each driving variable, all lags up to the model order are considered.

\begin{table}
    \footnotesize
    \centering
    \begin{tabular}{|c|cccccc|}
    \hline
         \backslashbox{$\tau$}{$k$}&1&	2&	3&	4&	5&	6\\
         	\hline\hline
        1&	\textbf{-0.172}&	0.007&	\textbf{-0.039}&	-0.005&	-0.007&	\textbf{-0.142}\\
        2&	\textbf{-0.083}&	\textbf{0.102}&	-0.007&	0.001&	\textbf{0.137}&	\textbf{-0.010}\\
        3&	\textbf{0.014}&	\textbf{0.078}&	\textbf{-0.012}&	0.006&	\textbf{-0.125}&	\textbf{-0.095}\\
        \hline

    \end{tabular}
    \caption{Weights of lagged variables for the prediction of the first target variable of VAR(P=3) model with 6 variables assigned by LAVARNET ($T$=3) after training. With bold we denote the 12 highest weights in absolute value, $k$ denotes the variable index and $\tau$ the lag.}
    \label{tab:interpretation_est}
\end{table}

\begin{table}
    \footnotesize
    \centering
    \begin{tabular}{|c|cccccc|}
    \hline
         \backslashbox{$\tau$}{$k$}&1&	2&	3&	4&	5&	6\\
         	\hline\hline
        1&1&1&0&0&1&1\\
        2&1&1&0&0&1&1\\
        3&1&1&0&0&1&1\\
        \hline

    \end{tabular}
    \caption{Lagged variables of VAR(P=3) model with 6 variables, where the lagged variables that drive the first target variable are indicated by 1, the rest lagged variables are indicated by 0, $k$ denotes the variable index and $\tau$ the lag.}
    \label{tab:interpretation_true}
\end{table}

In Table~\ref{tab:interpretation_est} an example case of weights assigned by LAVARNET to the lagged variables\footnote{Actually, the weights are assigned to the model's output vectors $y_{t,k}$ that correspond to certain lagged variables as stated in the model's description, but this is omitted here} for brevity. of a multivariate time series is presented. More precisely, LAVARNET is trained on forecasting future values of a multivariate time series (generated by the VAR(P=3) model and having $K=6$ variables) based on past values of all system variables and we present the weights of lagged variables that correspond to the first target variable. In Table~\ref{tab:interpretation_true} the true lagged variables that contribute to the evolution of the first target variable of the system, are shown. As one can see, 10 out of 12 important lagged variables are assigned as such by LAVARNET and also it produces 2 mismatches, giving $R_L$=83.33\% success percentage. Also, all 4 driving variables are correctly assigned as such, giving $R_V$=100\%.

\begin{figure}
    \centering
        \subfloat[]{\includegraphics[width=0.5\textwidth]{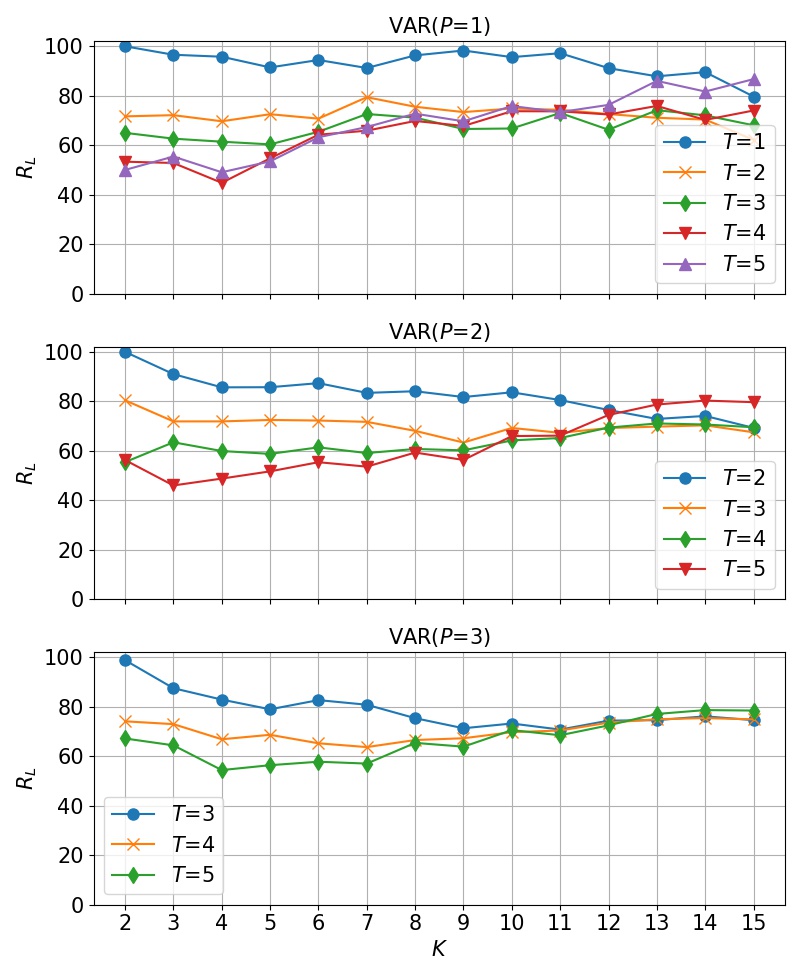}}
        \subfloat[]{\includegraphics[width=0.5\textwidth]{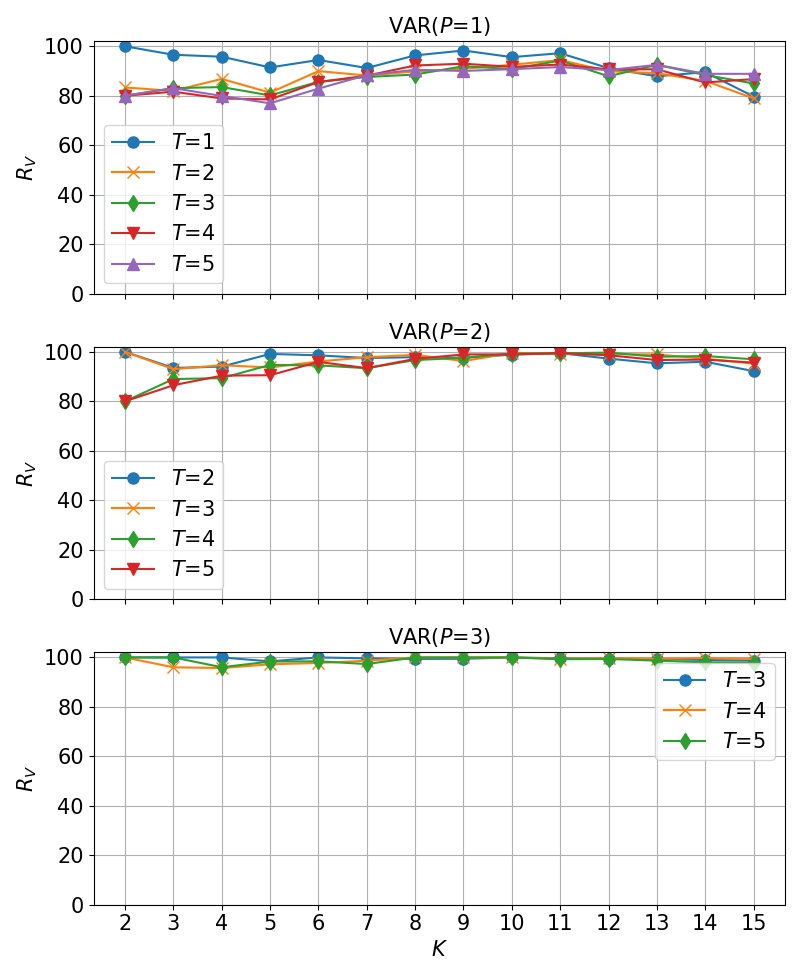}}
    \caption{The average success percentage for (a) lagged variables ($R_L$) and (b) variables ($R_V$), across 10 Monte-Carlo simulations using LAVARNET for different VAR order $P$, number of time steps $T$ and number of variables $K$.}
    \label{fig:interpretation}
\end{figure}

Aggregated results for all simulation scenarios and Monte-Carlo simulations are presented in Figure~\ref{fig:interpretation}. 
In the simulation scenarios, it is observed that LAVARNET's average success percentage in identifying the correct lagged variables is close to 70\%. While in the task of correctly identifying the driving variables it reaches values greater than 90\% and even 100\% in some cases. As expected, it is harder for LAVARNET to choose the correct lagged variables as the number of time steps $T$ increases. Additionally, the latter is easier as the VAR order $P$ increases, because the same variables contribute more intensely (with more lags). Interestingly, the number of variables $K$ does not seem to adversely affect the correct identification of important lagged variables as it increases. On the contrary, less variability in success percentages with respect to different time steps $T$ is exhibited as $K$ increases.

\subsection{Computational cost}
Except for the forecasting accuracy, another aspect of a model's performance is computational cost. In Table~\ref{tab:cost}, the average time required for model initialization and training on SML2010 data set, is illustrated for all neural network-based models of our study and for the same number of epochs (70).
Ten realizations are considered in order to present average performance. Also, one GPU device (GeForce GTX 1080) is employed for the computations.

The differences in execution time are not substantial among the competitive models, except RNN which is considerably faster and WaveNet which is considerably slower. However, RNN being fast comes as compensation for its poor forecasting accuracy. Among the models that perform well in forecasting LAVARNET is the fastest while FR-LAVARNET and R-LAVARNET follow right after leaving DARNN be the slowest.

\begin{table}
    \footnotesize
    \centering
    \begin{tabular}{l|l}
    \hline
         model& time \\
         \hline\hline
         LAVARNET&31.120 sec\\
         R-LAVARNET&40.290 sec\\
         FR-LAVARNET&33.796 sec\\
         LSTM&30.562 sec\\
         RNN&17.085 sec\\
         DARNN&41.166 sec\\
         WaveNet&101.865 sec\\
         \hline
    \end{tabular}
    \caption{The average time in seconds that it takes to form the graph and conduct the training on SML2010 data set per model.}
    \label{tab:cost}
\end{table}

\section{Conclusions}\label{sec:conclusions}
In this work, we propose a novel neural network architecture that leverages intrinsically estimated high dimensional latent representations of lagged variables, to make multivariate time series forecasts. This model is evaluated on one simulated data set and four real data sets from meteorology, music, solar activity, and finance areas and it is found to outperform other baseline and state of the art neural network architectures and machine learning models in most of the experiments. Moreover, its behavior is interpretable by the trainable weights' values it assigns to lagged variables as it is shown by a separate simulation study.

However, our architecture did not exhibit superior performance across half of the experiments on the data set from finance, on which it performed up to the mark though. The state of the art neural network DARNN exhibits great performance in the rest experiments conducted on this data set and also WaveNet produces accurate forecasts but not the best among this ensemble of models. In the cases that DARNN outperforms LAVARNET, information from the multivariate signals is better exploited by DARNN in terms of connectivity estimation, in terms of temporal modeling or both. A plausible explanation to this might be that LAVARNET considers a stable over time causality network of the underlying mechanism that generates the measurements and in finance slight relative variability (or noise) might occur. On one hand, this might seem like a limitation, on the other hand knowing it is useful twofold (a) the user considers using it on suitable data sets or/and (b) an expert splits the data set into relatively stable (in terms of who is driving who) periods and the model is applied separately. In conclusion, this should not be a problem to the vast majority of data sets as all coupled systems preserve their coupling structure, either for long or for short periods, and at every phase transition the model can be re-trained.

Finally, the conducted experiments indicate that recurrent neural networks (even the baselines) are more powerful in temporal modeling and especially time series forecasting than convolutional based architectures (WaveNet) which still produce accurate predictions though. A combination of convolutional layers and LAVARNET seems like a promising extension of the current model that the authors will consider as future work. Future work will also focus on improving the proposed architecture in the direction of reducing memory consumption and computational cost.

\section*{Acknowledgments}
This work is partially funded by the European Commission under the contract number H2020-761634 FuturePulse.

\bibliographystyle{elsarticle-num}
\bibliography{references}

\end{document}